\documentclass{article}

\PassOptionsToPackage{numbers, compress}{natbib}

\usepackage[preprint]{neurips_2021}

\usepackage{amsmath}
\usepackage[utf8]{inputenc} 
\usepackage[T1]{fontenc}    
\usepackage{hyperref}       
\usepackage{url}            
\usepackage{booktabs}       
\usepackage{amsfonts}       
\usepackage{nicefrac}       
\usepackage{microtype}      
\usepackage{xcolor}         

\usepackage{graphicx}
\usepackage{multirow}
\usepackage{amsthm}

\usepackage{multicol}
\usepackage{xpatch}
\usepackage{subcaption}
\usepackage{framed} 
\usepackage{amssymb}
\usepackage{algpseudocode}
\usepackage[ruled,vlined]{algorithm2e}
\usepackage{bbm}
\usepackage{comment}
\usepackage{array}
\newcolumntype{C}[1]{>{\centering\arraybackslash}p{#1}}

    \makeatletter
\def\@fnsymbol#1{\ensuremath{\ifcase#1\or \dagger\or \ddagger\or
   \mathsection\or \mathparagraph\or \|\or **\or \dagger\dagger
   \or \ddagger\ddagger \else\@ctrerr\fi}}
    \makeatother
    
\title{EDDA: Explanation-driven Data Augmentation to Improve Explanation Faithfulness}

\author{%
  Ruiwen Li \thanks{Equal contribution, co-first authors.}\\
  University of Toronto \\
  \texttt{ruiwen.li@mail.utoronto.ca} \\    
    \And
  Zhibo Zhang \footnotemark[1] \\
  University of Toronto\\
  \texttt{zhibozhang@cs.toronto.edu} \\
  \And
  Jiani Li  \\
  University of Toronto\\
  \texttt{nini.li@mail.utoronto.ca} \\
  \And
  Chiheb Trabelsi\\
  University of Toronto\\
  \texttt{chiheb.trabelsi@utoronto.ca}\\
  \And
  Scott Sanner  \\
  University of Toronto\\
  \texttt{ssanner@mie.utoronto.ca} \\
    \And
  Jongseong Jang\\
  LG AI Research\\
  \texttt{j.jang@lgresearch.ai} \\ 
  \And
  Yeonjeong Jeong\\
  LG AI Research\\
  \texttt{yj.jeong@lgresearch.ai} \\
  \And
  Dongsub Shim\\
  LG AI Research\\
  \texttt{dongsub.shim@lgresearch.ai} \\
}

\begin{document}

\maketitle

\begin{abstract}
Recent years have seen the introduction of a range of methods for post-hoc explainability of image classifier predictions. However, these post-hoc explanations may not always be faithful to classifier predictions, which poses a significant challenge when attempting to debug models based on such explanations. To this end, we seek a
methodology that can improve the faithfulness of an explanation method with respect to model predictions which does not require ground truth explanations.
We achieve this through a novel explanation-driven data augmentation (EDDA) technique that augments the training data with occlusions inferred from model explanations; this is based on the simple motivating principle that \emph{if} the explainer is faithful to the model \emph{then} occluding salient regions for the model prediction should decrease the model confidence in the prediction, while occluding non-salient regions should not change the prediction.  To verify that the proposed augmentation method has the potential to improve faithfulness, we evaluate EDDA using a variety of datasets and classification models. We demonstrate empirically that our approach leads to a significant increase of faithfulness, which can facilitate better debugging and successful deployment of image classification models in real-world applications. 
\end{abstract}

\textbf{Dual submission declaration}: This paper has been submitted to the main NeurIPS 2021 conference.

\section{Introduction}\label{introduction}

Deep learning models based on Convolutional Neural Networks(CNNs) have demonstrated superior performance in various computer vision tasks such as image classification~\cite{krizhevsky2009learning, he2016deep, simonyan2014very}, object detection~\cite{Redmon_2016_CVPR, NIPS2015_14bfa6bb}, and semantic segmentation~\cite{1411.4038, 1505.04597, 1606.00915}. However, the interpretability of these models has always been a concern as they often operate as black boxes. 
Therefore, researchers have recently explored many methodologies for explaining neural networks ~\cite{esteva2019guide, chattopadhay2018grad, covert2020understanding, chong2017deep, goyal2019counterfactual, smilkov2017smoothgrad, sundararajan2017axiomatic, lundberg2017unified, ribeiro2016should, chattopadhay2018grad, simonyan2013deep, devries2017improved, pruthi2020estimating, koh2017understanding, yeh2018representer}. A popular group of post-hoc explanation methods is saliency-based methods~\cite{simonyan2013deep, sundararajan2017axiomatic, smilkov2017smoothgrad, selvaraju2017grad}, which identify the salient regions of the input images through input gradients. 

With the development of explanation methodologies, some critical problems arise: What are the desired qualities of a good explainer?  And how can we improve the explanation methodologies to have these qualities? Existing works have investigated explanation robustness, where similar inputs should have similar explanations~\cite{alvarezmelis2018robustness,ghorbani2019intepretation, yeh2019fidelity}. In this paper, we explore another desired quality of explanation methodologies: faithfulness. Whereas many existing works have explored the idea of faithfulness~\cite{yeh2019fidelity, sippy2020data, jacovi2020towards, lakkaraju2019faithful}, there is no universal agreement on both its definition and evaluation. In this paper, we believe that a faithful explanation (salient region) related to an input  
image should contain the crucial information that the classifier utilizes to make the prediction. Therefore, the model should be able to recognize the input with high confidence (we consider here the confidence score as being the softmax probability related to the class predicted by the model \cite{sattarzadeh2020explaining} 
after masking the unimportant region according to a faithful explanation (see equation (\ref{drop_increase_eq}))).

\begin{figure}[t!]
    \centering
    \begin{subfigure}[b]{1\linewidth}
    \centering
        \includegraphics[width=0.18\linewidth]{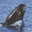}
        \includegraphics[width=0.18\linewidth]{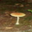}
        \includegraphics[width=0.18\linewidth]{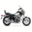}
        \includegraphics[width=0.18\linewidth]{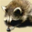}
        \includegraphics[width=0.18\linewidth]{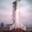}
        \caption{Vanilla images}
    \end{subfigure}

    \begin{subfigure}[b]{1\linewidth}
        \centering\includegraphics[width=0.18\linewidth]{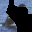}
        \centering\includegraphics[width=0.18\linewidth]{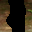}
        \centering\includegraphics[width=0.18\linewidth]{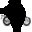}
        \centering\includegraphics[width=0.18\linewidth]{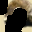}
        \centering\includegraphics[width=0.18\linewidth]{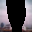}
        \caption{Masking under the classifier trained with \textbf{no data augmentation}}
    \end{subfigure}

    \begin{subfigure}[b]{1\linewidth}
        \centering\includegraphics[width=0.18\linewidth]{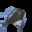}
        \centering\includegraphics[width=0.18\linewidth]{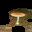}
        \centering\includegraphics[width=0.18\linewidth]{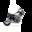}
        \centering\includegraphics[width=0.18\linewidth]{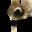}
        \centering\includegraphics[width=0.18\linewidth]{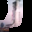}
        \caption{Masking under the classifier trained with \textbf{EDDA}}
    \end{subfigure}
    \caption{Example of unfaithful and faithful explanations. (a) The original training images with class whale, mushroom, motorcycle, raccoon and rocket from left to right. (b)The classifier is trained with no data augmentation. After masking out the 50\% unimportant regions given by GradCAM, the predicted classes changed into Pine tree, lamp, lamp, lizard and wardrobe accordingly. (c) The classifier is trained with EDDA. After masking out the 50\% unimportant regions given by GradCAM, the predicted classes remain to be whale, mushroom, motorcycle, raccoon and rocket accordingly. With EDDA, the explainer successfully identify salient regions that contain the important information for classification.}
    \label{fig:example}
\end{figure}
While providing interpretations of the predictions, the explanations generated by the saliency-based explainers may not always be faithful to the model in question. For example, as shown in Figure \ref{fig:example}, after masking the unimportant regions provided by the explainer, the model trained with no augmentation fails to predict the desired label of the image, indicating that the "salient region" provided by the explainer is lacking crucial information for a correct
classification. Such cases can easily confuse human users and thus harm the trustworthiness of the model.

In this paper, we aim to improve the faithfulness aspect of the explanations with respect to the model predictions. To accomplish this,
we propose explanation-driven data augmentation (EDDA) that augments the training data with occlusions of existing data stemming from model-explanations. Specifically, we generate two occlusions of each input image, namely \textit{umask} (unimportant-mask) that occludes unimportant/unsalient region and \textit{imask} (important-mask) that occludes important/salient region. In principle, the prediction of \textit{umask} images should be the same as the original image and  the prediction of \textit{imask} images should be different. A similar rationale is considered for natural language processing models in \cite{deyoung2019eraser} where preserving and removing the important parts of the text should respectively keep and alter the model's decisions accordingly.
To verify the effectiveness of our proposed methodology, we conduct experiments on two image datasets (CIFAR-100 \cite{krizhevsky2009learning} and PASCAL VOC 2012 \cite{pascal-voc-2012}) using two CNN ( ResNet~\cite{he2016deep} and VGG~\cite{simonyan2014very} ) models with GradCAM~\cite{selvaraju2017grad} as the explainer. We empirically demonstrate that EDDA yields more faithful explanations across different datasets and models 
compared to the non-explanation-driven data augmentation methods.

\section{Related Work}

\paragraph{Post-hoc Explanations} Post-hoc methods explain the decisions made by a pre-built black-box model.  Most of the current post-hoc explanation methods focus on local explanations, which describe the model behavior at the neighborhood of an individual prediction. Early post-hoc explanation methods, such as LIME \cite{ribeiro2016should} and SHAP \cite{lundberg2017unified}, attribute importance scores to input features with respect to the prediction of the target data point. Gradient-based methods are broadly used in explaining Convolutional Neural Networks (CNNs): Saliency Map \cite{simonyan2013deep} utilizes the input gradient of the network to highlight the critical pixels for a prediction; Integrated Gradients \cite{sundararajan2017axiomatic} addresses the gradient saturation issue by generating interpolants from the baseline to the input image and summing over the gradient of each interpolant; SmoothGrad \cite{smilkov2017smoothgrad} addresses the same issue, and it computes sharper saliency maps by adding noise to copies of the input image and averaging the input gradient of these copies. The Class Activation Map (CAM) \cite{zhou2016learning} based explxanation methods are commonly used in the literature. Grad-CAM \cite{selvaraju2017grad} employs the weighted activations of the feature maps in the final convolutional layer, where the weights are computed based on the gradient of an output class with respect to each feature map. Despite the variety of these methods, they were either evaluated qualitatively or based on ground truth localization information such as object detection bounding boxes. It may not be meaningful if the dataset is biased, e.g., the classifier can utilize the sea background of the fish objects or the sky background of the plane object to make decisions that overfits specific training data. Recent image-based explanation methods such as Grad-CAM++ \cite{chattopadhay2018grad} and SISE \cite{sattarzadeh2020explaining} adopt the Drop\% and Increase\% metrics in the performance evaluation. These metrics capture aspects of faithfulness that we also consider in our work (see equation (\ref{drop_increase_eq})).

\paragraph{Data Augmentation} Data augmentation is a training strategy that improves the performance of deep learning models by applying various transformations to the original training data. These transformations are relatively simple but effective in increasing the data diversity and the model's generalization ability. Cutout \cite{devries2017improved} is a regional dropout method that zero-masks a random fix-sized region of each input training image to improve the model's test accuracy. MixUp \cite{zhang2017mixup} linearly combines two training inputs where their targets are linearly interpolated in the same fashion. CutMix \cite{yun2019cutmix} builds on Cutout, and it avoids the information loss in the dropout region by randomly removing and mixing patches among training images where the target labels assigned are proportional to the size of those patches. These methods show improved test accuracy, but they have never been verified from the perspective of explanation. Besides, some methods aim to improve model robustness and uncertainty \cite{hendrycks2020augmix}. In this paper, we provide a novel augmentation method that improves the potential of
explanation faithfulness and allows one to perform sanity checks
by conducting stress tests. This sanity check is critical as it permits to verify whether the accuracy improvement benefits from the bias in the dataset or not.

\paragraph{Explanation faithfulness} Existing works have studied the concept of explanation faithfulness (fidelity) from the following perspectives:
\begin{itemize}
    \item In \cite{guidotti2019survey}, the authors define fidelity as how much the explanation model mimics the behavior of a black-box in terms of prediction score~\cite{jacovi2020towards}.
    \item An alternative definition of fidelity~\cite{yeh2019fidelity} is the expected difference between the dot product
of the input perturbation to the explanation and the output perturbation.
    \item Additionally, data staining~\cite{sippy2020data} is also introduced as an evaluation method of explanation faithfulness. It trains a stained predictor (i.e. a model that is biased to err systematically) and evaluates the explainer’s ability to recover the stain.
\end{itemize}
However, these methods do not directly relate to our work as it either requires the generation of random input perturbations or the retraining of a separate model. Inspired by the work in \cite{deyoung2019eraser}, we present in section \ref{propose}  a way to measure faithfulness.

\section{Proposed Method}\label{propose}

In this section, we describe the proposed explainer-driven data augmentation (EDDA) method in detail.

We begin by clarifying explanation faithfulness. the authors in \cite{deyoung2019eraser} have presented faithfulness from the perspectives of comprehensiveness and sufficiency. Comprehensiveness stands for the difference between the confidence score for the vanilla input and the confidence score for the perturbed input that preserves the unimportant region. In this context, a confidence score is the softmax probability for the predicted class. Sufficiency stands for the confidence score difference between the vanilla input and the perturbed input that reserves the important region. In this paper, we are going to adopt the sufficiency aspect for explanation faithfulness and use it as a measurement in the experimental part, i.e.,
\begin{equation}
    \mathit{faithfulness}_{\mathit{sufficiency}} (\phi, \hat{y_i}, x_i) := \phi(x_{i,umask})^{\hat{y_i}} - \phi(x_i)^{\hat{y_i}}
\end{equation}
where $x_i$ is the original input image. $\hat{y_i}$ is the predicted class that we want to explain the decision with respect to. \textit{umask} images mask out the unimportant regions of the input images based on the explanation scores of the input images. For any input $x$, $\phi(x)^{c}$ produces the confidence score of it with respect to the class label $c$. Ideally, a faithful explainer to the model should observe either a large increase or a small decrease in the prediction confidence of the \textit{umask} image, where both cases imply a larger sufficiency score.

\subsection{Motivation}

As stated in the previous section, with a faithful explanation, the model should still be able to recognize the image as the original label when occluding the unimportant regions. In contrast, the predicted label should be different when occluding important regions. One way to encourage this behaviour is augmenting the training data. Specifically, for each training input, we generate two masked images, namely \textit{umask} (unimportant-mask) and \textit{imask} (important-mask). The \textit{umask} images mask out the unimportant regions and the \textit{imask} images mask out the important regions. If the prediction of the \textit{imask} image is  correct, it suggests that the explanation may not be faithful to the prediction, and we train on the \textit{imask} image. By doing this, we force the model to focus on some other region and we may obtain a faithful explanation next time. If the prediction of the \textit{umask} image is  correct, it suggests that the explanation is faithful, and we train on the \textit{umask} image to reinforce this faithfulness.

Motivated by the idea above, we propose our explainer-driven data augmentation method. 

\subsection{EDDA for Multi-class Classification}

\begin{figure}[ht]
    \centering
    \includegraphics[width=0.99\linewidth]{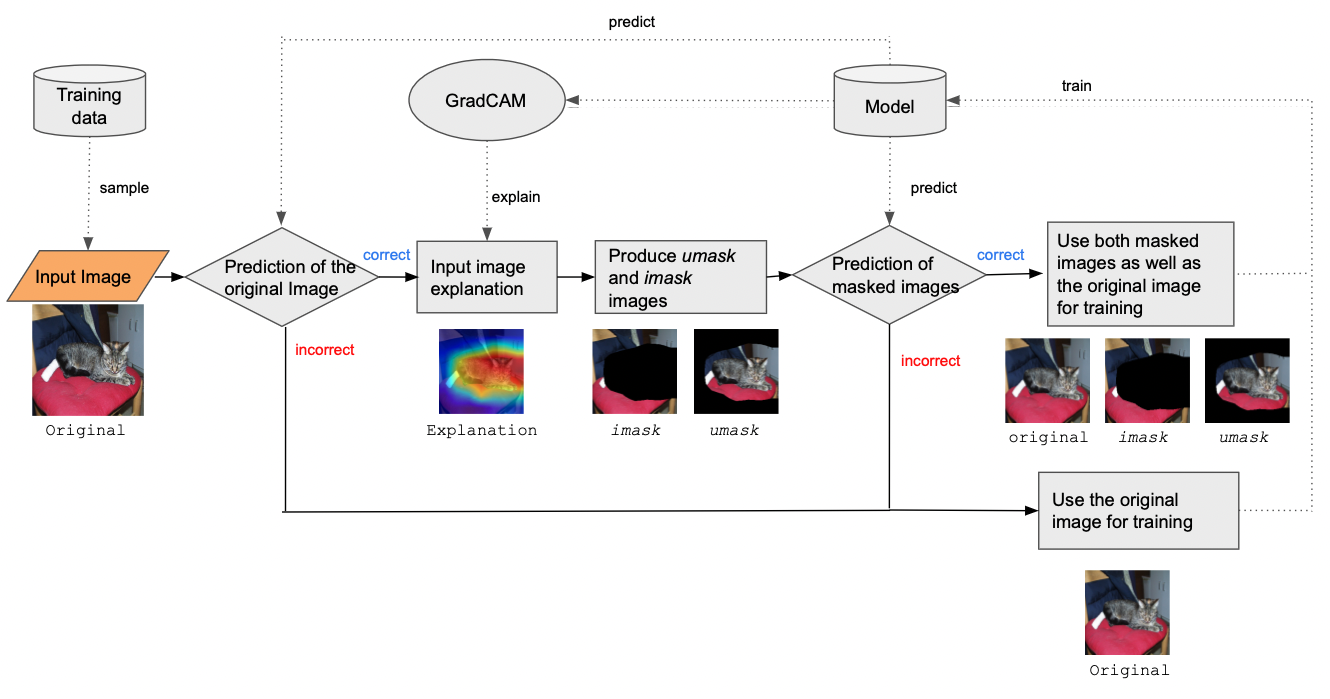}
    \caption{Explainer-driven data augmentation for multi-class image classification}
    \label{fig:method}
\end{figure}

Figure \ref{fig:method} shows the training process of EDDA in the multi-class classification setting. Let $\mathcal{M}$ denotes the training model, then in each epoch, for an input image $x \in \mathbb{R}^{w \times h \times c}$ with a predicted label $y = \mathcal{M}(x)$ and a ground truth label $\hat{y}$, we only perform augmentation when the prediction is correct, i.e. $y=\hat{y}$. Otherwise, we train the classifier with the original image alone.

If the prediction is correct, we then use the explainer $f_{\mathcal{M}}$ to obtain the saliency map $s_{x}^{y} $ indicating the attribution of each pixel to the prediction $y$:
\begin{equation}
\label{eq:1}
    s_{x}^{y} = f_{\mathcal{M}}(x, y)
\end{equation}
where $s_{x}^{y} \in \mathbb{R}^{w \times h}$ and is normalized into the range of $[0, 1]$. 

Next, we generate two masked images. We generate the \textit{umask} image $x_{umask}$ by preserving the pixels with saliency scores that are above a threshold $\tau_{umask}$ (a hyperparameter) and masking out other pixels:
\begin{equation}
\label{eq:2}
    x_{umask} = x \odot (s_{x}^{y}  > \tau_{umask}),
\end{equation}

where $x_{umask} \in \mathbb{R}^{w \times h \times c}$. Correspondingly, we generate \textit{imask} image $ x_{imask}$ by occluding the pixels with saliency scores above a threshold $\tau_{imask}$:
\begin{equation}
\label{eq:3}
    x_{imask} = x \odot (s_{x}^{y}  < \tau_{imask}).
\end{equation}

We obtain the prediction $\hat{y}_{umask}$ and $\hat{y}_{imask}$ on the perturbed images $x_{umask}$ and $x_{imask}$ by feeding them into the classifier $\mathcal{M}$. 

If the \textit{imask} image has the same prediction as the original image (i.e. $\hat{y}_{imask} = y$), meaning that the explanation may not be faithful and there is still discriminative information in the masked image, we then label the \textit{imask} image as the original image's prediction label $y$, and add the data $(x_{imask}, y)$ into the current training batch. If the \textit{imask} images has a different prediction than the one for the original image, we do not include the \textit{imask} image for training.

Similarly, if the \textit{umask} image has the same prediction as the original image (i.e. $\hat{y}_{umask} = y$), meaning that the explanation is faithful and the \textit{umask} image preserves key information for classification, we then add the \textit{umask} image with the image's prediction label $(x_{umask}, y)$ into the current training batch. Otherwise, we do not include the \textit{umask} image in the training process. We summarize our proposed pipeline in Algorithm \ref{algor1} in the appendix.

\subsection{EDDA for Multi-label Classification}

 We propose a separate method in the multi-label classification setting, where there are multiple ground truth labels for a single image. This raises the needs for doing independent explanations with respect to different labels. We first make predictions on the input image, and then select the set of classes that have positive predictions and align with the ground truth labels - the \texttt{True-Positive} classes. We explain the input image with respect to each of these classes and generate the according \textit{imask} images. For each \textit{imask} image, if the prediction is correct on the according label, then we add that \textit{imask} image with the original labels into the current training batch. The full algorithm is suggested in Algorithm \ref{algor2}.

\section{Experiments and Evaluations}
\label{sec:experiments}

We perform a range of experiments with multiple datasets and models to evaluate the performance of EDDA against other methodologies. The experiment demonstrates the effectiveness of EDDA on the improvement of explanation faithfulness. We further investigate the visualization of masked images along the training process of EDDA and show that explanations are improved over time.
\subsection{Experiment Settings}
\paragraph{Datasets}
We use CIFAR-100 \cite{krizhevsky2009learning} for multi-class classification and  PASCAL VOC 2012 \cite{pascal-voc-2012} for multi-label classification. Details about train and validation split and data pre-processing are stated in the appendix.

\paragraph{Models}
To verify our consistent performance on different backbone models, we performed experiments on two popular convolution networks: ResNet-50 \cite{he2016deep} and VGG-16 \cite{simonyan2014very}.

\paragraph{Hyperparameters}
We tuned several method-based hyperparameters, including the threshold parameters $\tau_{imask}$ and $\tau_{umask}$in EDDA, as well as the Beta distribution parameter \emph{$\alpha$} in MixUp \cite{zhang2017mixup}. For fair comparison, we used the same training hyperparameters settings (e.g. batch size, number of epochs) in all methods. Details are in the appendix.

\subsection{Experiment Methods}
\paragraph{Baselines}
For the purpose of thoroughly verifying the explanation capability of the EDDA method, we implemented the following non-explanation driven baselines as comparisons.
\begin{itemize}
\item \textbf{No Augmentation}: the vanilla training pipeline without any forms of data augmentation.
\item \textbf{CutMix} \cite{yun2019cutmix}: it belongs to a type of regional dropout method, which randomly removes some image regions and fills them with patches from other training images. The target labels are assigned based on the proportion of the area of those patches.
\item \textbf{MixUp} \cite{zhang2017mixup}: it does not use regional drop but linearly blends two training inputs instead. The targets of the augmented images are assigned based on the blending ratio of their source images.
\item \textbf{AugMix} \cite{hendrycks2020augmix}: it creates several augmentations of every input and then linearly combine them with the original images by sampled weights. The targets remain unchanged. Similar to the original paper, we adopted the Jensen-Shannon Consistency Loss. \end{itemize}

\paragraph{Our Methods}
In terms of the explanation-driven data augmentation, we performed experiments under the EDDA framework with the following two explainers:
\begin{itemize}
\item \textbf{Random Explainer (RandExp)}: it attributes a random saliency score to every pixel uniformly in the range of $[0, 1]$. It is a trivial but important baseline.
\item \textbf{GradCAM (for EDDA)}:
the explainer we chose to demonstrate EDDA's capability. It uses the gradients of any target concept flowing into the final convolutional layer to produce a coarse localization map highlighting the important regions in the image for predicting the concept \cite{selvaraju2017grad}. 
\end{itemize}

\paragraph{Evaluation Metrics}

\label{sec:metrcis}
We applied the metrics of Drop\% and Increase\% introduced in \cite{ramaswamy2020ablation, fu2020axiom, sattarzadeh2020explaining} which can exactly demonstrate our claim. Drop\% measures the positive attribution loss after a \emph{umask}, while Increase\% measures the negative attribution discard \cite{sattarzadeh2020explaining}. We adapted these two metrics, where for both Drop\% and Increase\% we measure the average magnitude of drop and increase of the confidence score with respect to the predicted label after occluding least salient regions, which is defined as

    \begin{equation}\label{drop_increase_eq}
    \begin{split}
    Drop\% = \frac{1}{N}\sum_{i=1}^{N}\frac{\mathrm{max}(0, \phi(x_i)^{\hat{y_i}} - \phi(x_{i,umask})^{\hat{y_i}})}{\phi(x_i)^{\hat{y_i}}} \times 100\\
    Increase\% = \frac{1}{N}\sum_{i=1}^{N}\frac{\mathrm{max}(0, \phi(x_{i,umask})^{\hat{y_i}} - \phi(x_i)^{\hat{y_i}})}{\phi(x_i)^{\hat{y_i}}} \times 100\\
    \end{split}
    \end{equation}
    where $x_i$ represents the input image, $\mathcal{M}$ represents the classifier and returns a predicted label, $\phi(x_i)^{\hat{y_i}}$ gives a confidence score of input $x_i$ with respect to $\hat{y_i}$, where $\hat{y_i}=\mathcal{M}(x_i)$ is the predicted label.

In our experiments, we set the top 15\% as the threshold (following previous work \cite{sattarzadeh2020explaining}) for selecting the most salient pixels for reservation. We also tried other thresholds including 30\% and 50\% and they displayed similar trends. If the explainer is faithful to the model, the \emph{umasked} image should preserve the most discriminative regions \cite{ramaswamy2020ablation}, hence has a high confidence score which leads to a lower Drop\% and a higher Increase\%. These metrics do not require ground truth bounding boxes for the classified objects, which are expensive to label especially in real-world applications.

\subsection{Performance Comparison on Explanation Faithfullness}
\begin{table}[ht]
\caption{ResNet-50 and VGG-16 model performance in terms of Drop\%\textsubscript{mag} and Increase\%\textsubscript{mag} on CIFAR-100 and PASACL VOC 2012. 
}
    \centering
    
\begin{subtable}{\textwidth}
 \begin{tabular}{ |p{1.26cm}|p{2.7cm}|p{2.7cm}|p{2.7cm}|p{2.7cm}|  }
 \hline
    Model & \multicolumn{2}{|c|}{ResNet-50}    &\multicolumn{2}{|c|}{VGG-16} \\
 \hline
    Metric&  \multicolumn{1}{|C{2.4cm}|}{$Drop\%$ \newline (Lower is better)}  &  \multicolumn{1}{|C{2.4cm}|}{$Increase\%$ \newline (Higher is better)}   & \multicolumn{1}{|C{2.4cm}|}{$Drop\%$ \newline (Lower is better)}  &\multicolumn{1}{|C{2.4cm}|}{$Increase\%$ \newline (Higher is better)}  \\
 \hline
  No Aug &$96.15 \pm 0.30$ &$0.35 \pm 0.09$ &$93.41 \pm 0.52$ &$0.25 \pm 0.06$ \\
  CutMix &$96.38 \pm 0.34$ &$0.24 \pm 0.02$ &$95.09 \pm 0.36$ &$0.24 \pm 0.04$ \\
  MixUp &$94.80 \pm 0.47$ &$0.62 \pm 0.15$ &$93.80 \pm 0.33$ &\textbf{0.99} $\pm$ \textbf{0.23}  \\
  AugMix &$95.86 \pm 0.17$ &$0.32 \pm 0.03$ &$94.29 \pm 0.55$ &$0.22 \pm 0.06$ \\
  RandExp &$95.91 \pm 0.31$ &$0.39 \pm 0.05$ &$93.99 \pm 0.57$ &$0.33 \pm 0.11$ \\
  EDDA &\textbf{82.22} $\pm$ \textbf{0.30} &\textbf{1.32} $\pm$ \textbf{0.13} &\textbf{88.15} $\pm$ \textbf{0.60} &$0.45 \pm 0.12$ \\
 \hline
\end{tabular}
    \caption{CIFAR-100}
 \end{subtable}

\begin{subtable}{\textwidth}
 \begin{tabular}{ |p{1.26cm}|p{2.7cm}|p{2.7cm}|p{2.7cm}|p{2.7cm}|  }
 \hline
    Model & \multicolumn{2}{|c|}{ResNet-50}    &\multicolumn{2}{|c|}{VGG-16} \\
 \hline
    Metric&  \multicolumn{1}{|C{2.4cm}|}{$Drop\%$ \newline (Lower is better)}  &  \multicolumn{1}{|C{2.4cm}|}{$Increase\%$ \newline (Higher is better)}   & \multicolumn{1}{|C{2.4cm}|}{$Drop\%$ \newline (Lower is better)}  &\multicolumn{1}{|C{2.4cm}|}{$Increase\%$ \newline (Higher is better)}  \\
 \hline
  No Aug &$53.84 \pm 3.95$ &$1.27 \pm 0.42$ &$40.71 \pm 2.30$ &$3.25 \pm 0.84$ \\
  CutMix &$62.52 \pm 4.90$ &$1.84 \pm 0.52$ &$48.55 \pm 2.34$ &$1.85 \pm 0.16$ \\
  MixUp &$55.67 \pm 1.28$ &$2.87 \pm 0.43$ &$46.83 \pm 0.94$ &$1.95 \pm 0.17$  \\
  AugMix &$49.02 \pm 3.21$ &$1.91 \pm 0.18$ &$41.10 \pm 1.97$ &\textbf{19.33} $\pm$ \textbf{13.54} \\
  RandExp &$75.39 \pm 5.76$ &$8.61 \pm 4.27$ &$45.28 \pm 1.23$ &$7.21 \pm 4.12$ \\
  EDDA &\textbf{42.72} $\pm$ \textbf{7.50} &\textbf{11.89} $\pm$ \textbf{14.57} &\textbf{9.01} $\pm$ \textbf{3.35} &$9.00 \pm 3.28$ \\
 \hline
\end{tabular}
    \caption{PASCAL VOC 2012}
 \end{subtable}
\label{tab:dropmag}
\end{table}
Table \ref{tab:dropmag} shows the Drop\% and the Increase\% evaluation results on the CIFAR-100 dataset and the PASCAL VOC 2012 dataset with both ResNet-50 and VGG-16 models. We report the average scores and 95\% confidence intervals over five independent runs with different random seeds.

It can be observed that EDDA has the best $Drop\%$ scores compared to all the baseline methodologies, especially with PASCAL VOC 2012 on VGG (30\% lower than others) . For $Increase\%$, EDDA peforms the best across both datasets with the ResNet model. Whereas EDDA ranks second on $Increase\%$ of VGG-16 model for both datasets, it still performs better than four baselines.

\begin{figure}[ht]
    \centering
    \includegraphics[width=1\linewidth]{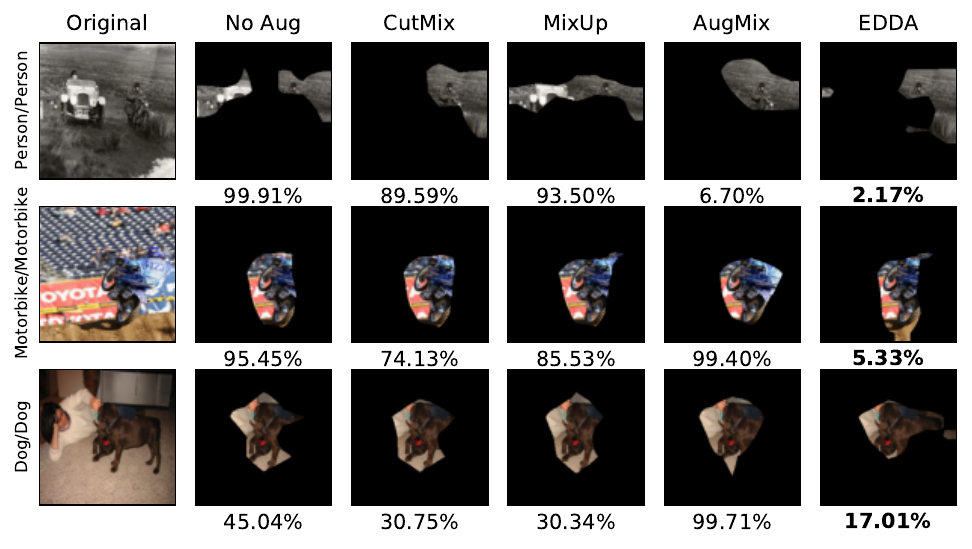}
    \caption{Sample test \textit{umask} images generated with different augmentation methods from  PASACL VOC 2012 dataset. The numbers below are the $Drop\%$ score and the title on the left is the target/prediction labels of each sample. All images are predicted correctly.}
    \label{fig:drop}
\end{figure}

To further understand the explainer behaviour with different augmentation techniques, we visualize the \textit{umask} images in Figure \ref{fig:drop}. Here, we have the following observations:
\begin{itemize}
    \item Quantitatively, $Drop \%$ scores of EDDA are significantly smaller than other methods.
    \item Qualitatively, EDDA generates \textit{umask} images that correspond more closely to the prediction labels. For example, in the Person sample in Figure~\ref{fig:drop}, the baselines generate \textit{umask} images either containing parts of the car (No Aug, MixUp) or parts of the bicycle(CutMix, AugMix), whereas EDDA only shows regions of the person.
\end{itemize}

\subsection{Explanation Improvement along the Training Process}
\begin{figure}[ht]
    \centering
    \includegraphics[width=0.8\linewidth]{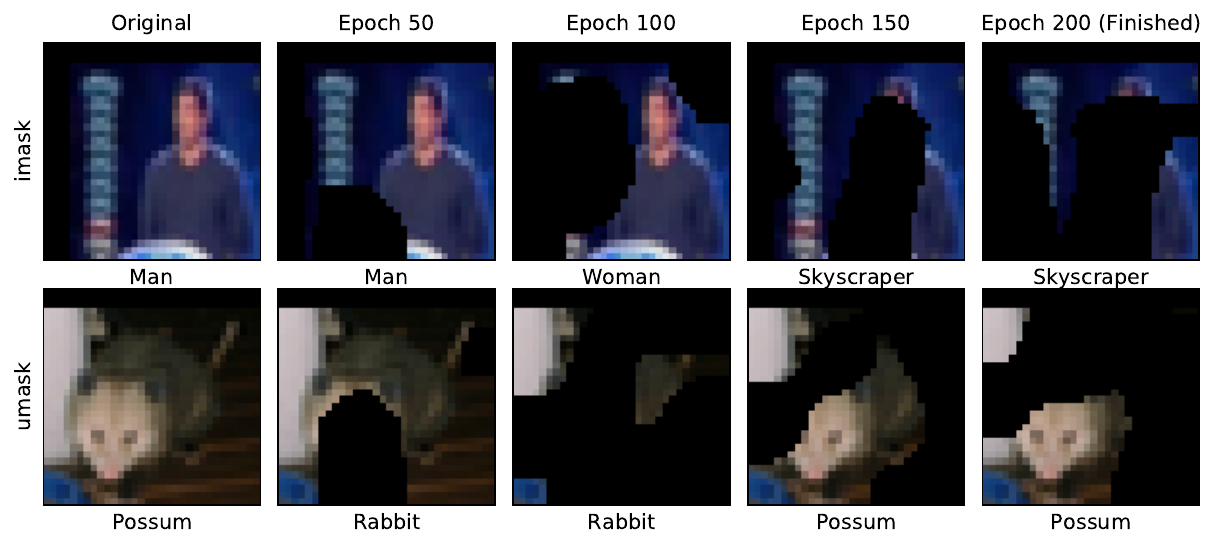}
    \caption{Sample \textit{imask} and \textit{umask} images generated at different epochs with EDDA from CIFAR-100 dataset on Resnet-50. The label below is the predicted label of each image}
    \label{fig:improving}
\end{figure}

To investigate the explanation faithfulness improvement of EDDA models along the growth of training iterations, we plot  \textit{imask} and \textit{umask} images ($\tau_{imask} = \tau_{umask} = 0.5$) generated at 4 model checkpoints (epoch 50, 100, 150 and 200) in Figure~\ref{fig:improving}. Here, we make the following key observations:
\begin{itemize}
    \item The predicted labels of the masked images change towards the desired direction along the training process. For the first example, the \textit{imask} image is still predicted as the original label "Man" at epoch 50 and a similar label "Woman" at epoch 100. At epoch 150 and 200, the prediction changes to "Skyscraper" that is totally different from the original prediction. For the second example, along the training process, the prediction of \textit{imask} images changes from a different label "Rabbit" to the original predicted label "Possum".
    
    \item Visually, the masked images identify the important/unimportant regions better along the process. Specifically, the \textit{imask} images contain less information associated with the original label "Man" in later epochs, whereas the \textit{umask} images contain more important information of "Possum" in later epochs. 
\end{itemize}

\section{Conclusion}

In this paper, we propose a novel explanation-driven data augmentation (EDDA) to improve the faithfulness of explanation methodologies with respect to the CNN model predictions. We verify our proposed method with two image datasets ( CIFAR-100 \cite{krizhevsky2009learning} and PASCAL VOC 2012 \cite{pascal-voc-2012}) and two CNN ( ResNet \cite{he2016deep} and VGG \cite{simonyan2014very} ) models with GradCAM \cite{selvaraju2017grad} as the explainer. Through these experiments, we empirically show that our method gives explanations that are more faithful to the model predictions across different datasets and CNN models both qualitatively and quantitatively compared to the non-explanation-driven augmentation methods and as well as no augmentation.

\bibliography{main}
\bibliographystyle{unsrt}

\newpage

\appendix

\section{Algorithms}
This section contains the algorithms that we use in the main sections of the paper.

 \begin{frame}{}
    \begin{algorithm}[H]
\SetAlgoLined
     \textbf{Input}: train set $\mathcal{D}$, classifier $\mathcal{M}$, explainer $f_{\mathcal{M}}(\cdot)$, training epochs $E$, mini-batch size $N$
     \\
     
    \For{$e \in \{1,...,E\}$}{
    
    \For{$(\mathcal{X}, \mathcal{Y}) \stackrel{N}{\sim} \mathcal{D}$}{

        $\Tilde{\mathcal{X}}, \Tilde{\mathcal{Y}} \leftarrow$ $\textbf{EDDA}(\mathcal{X}, \mathcal{Y}, \mathcal{M}, f_{\mathcal{M}})$ \Comment{\small use Algorithm \ref{algor1} or Algorithm \ref{algor2} depending on the task}
        
         $\mathcal{M}\leftarrow \mathrm{Train(\mathcal{M}, \Tilde{\mathcal{X}}, \Tilde{\mathcal{Y}})}$
         
    }

 }
 \caption{Training Pipeline for Explanation-driven Data Augmentation}
\end{algorithm}
\end{frame}{}

\begin{frame}{}
 \begin{algorithm}

\SetAlgoLined
     \textbf{Input}: input image batch $\mathcal{X}$, target label batch $\mathcal{Y}$, model $\mathcal{M}$, explainer $f_{\mathcal{M}}(\cdot)$, mini-batch size $N$
     \\
        $\mathcal{X}', \mathcal{Y}' \leftarrow \{\}, \{\}$\\
        // Get the attribution map using explainer $f_{\mathcal{M}}$\\
            $\mathcal{S}_{\mathcal{X}}^{\mathcal{Y}} \leftarrow f_{\mathcal{M}}(\mathcal{X}, \mathcal{Y})$ \Comment{Equation (\ref{eq:1})}
        
    // Mask the most salient regions using a hyperparameter $\tau_{imask}$, where \emph{imask} stands for important-mask\\
     $
            \mathcal{X}_{imask} \leftarrow \mathcal{X} \odot (\mathcal{S}_{\mathcal{X}}^{\mathcal{Y}} > \tau_{imask})
        $ \Comment{Equation (\ref{eq:2})}
    
    // Mask the most unsalient regions using a hyperparameter $\tau_{umask}$, where \emph{umask} stands for unimportant-mask\\
     $
            \mathcal{X}_{umask} \leftarrow \mathcal{X} \odot (\mathcal{S}_{\mathcal{X}}^{\mathcal{Y}} < \tau_{umask})
        $ \Comment{Equation (\ref{eq:3})}
        
        // Predict the labels based on \emph{imasked} and \emph{umasked} images.\\
        $
            \hat{\mathcal{Y}}_{imask} \leftarrow \mathcal{M}(\mathcal{X}_{imask})
        $\\
        $
            \hat{\mathcal{Y}}_{umask} \leftarrow \mathcal{M}(\mathcal{X}_{umask})
        $\\
        
        // Predict the labels based on original images.\\
        $\hat{\mathcal{Y}} \leftarrow \mathcal{M}(\mathcal{X})$
        
        \For{$1 \leq k \leq N$}{
 
          \If{$\mathcal{Y}[k] = \hat{\mathcal{Y}}[k]$}{
          \If{$\mathcal{Y}[k] = \hat{\mathcal{Y}}_{imask}[k]$}{
          // If the prediction on the imasked image aligns with the ground truth label, then\\
          // concatenate the imasked image and the original label into the current training batch.\\
          $\mathcal{X}' \leftarrow \mathcal{X}'\cup \{\mathcal{X}_{imask}[k]\}$\\
          $\mathcal{Y}' \leftarrow \mathcal{Y}'\cup \{\mathcal{Y}[k]\}$
          }
          
          \If{$\mathcal{Y}[k] = \hat{\mathcal{Y}}_{umask}[k]$}{
          // If the prediction on the umasked image aligns with the ground truth label, then\\
          // concatenate the umasked image and the original label into the current training\\
          // batch.\\
          $\mathcal{X}' \leftarrow \mathcal{X}'\cup \{\mathcal{X}_{umask}[k]\}$\\
          $\mathcal{Y}' \leftarrow \mathcal{Y}'\cup \{\mathcal{Y}[k]\}$
          }}
        }
\Return $\mathcal{X}'$, $\mathcal{Y}'$
 \caption{$\textbf{EDDA}_{mc}$: Explanation-driven Data Augmentation for Multi-class Classification}
\label{algor1}
 \end{algorithm}
\end{frame}{}

 \begin{frame}{}
    \begin{algorithm}[H]
\SetAlgoLined
     \textbf{Input}: input image batch $\mathcal{X}$, target label batch $\mathcal{Y}$, model $\mathcal{M}$, explainer $f_{\mathcal{M}}(\cdot)$\\
    $\mathcal{X}', \mathcal{Y}' \leftarrow \mathcal{X}, \mathcal{Y}$\\
    \For{$1 \leq k \leq |\mathcal{X}|$}{
        // Obtain the prediction given by the classifier.\\
        $
            \hat{\mathcal{Y}} \leftarrow \mathcal{M}(\mathcal{X}[k])
        $\\
        // Comparing with the ground truth labels, we select the \texttt{True-Positive} classes (where the \\
        // prediction and the label are both \texttt{True}) and denote the set to be $c$  \\
        $\mathcal{C} \leftarrow indices(\hat{\mathcal{Y}} = \mathcal{Y}[k])$
           
                // Get the importance maps for all correct labels at the same time using explainer $f_{\mathcal{M}}$.\\
        $
            \mathcal{S}_{\mathcal{X}}^{\mathcal{Y}} \leftarrow f_{\mathcal{M}}(\mathcal{X}[k], \mathcal{C})
        $
        
        // Mask the most salient regions using a pre-defined hyperparameter $\tau$.\\
        $
            \mathcal{\mathcal{X}}_{imask} \leftarrow \mathcal{X}[k] \odot (\mathcal{S}_{\mathcal{X}}^{\mathcal{Y}} > \tau)
        $
        
        // Predict the labels based on masked images.\\
        $
            \hat{\mathcal{Y}}_{imask} \leftarrow \mathcal{M}(\mathcal{\mathcal{X}}_{imask})
        $
        
        \For{$1\leq z \leq \mathbf{|\mathcal{C}|}$}{
            \If{$\hat{\mathcal{Y}}_{imask}[z] = \mathcal{Y}[k, z]$}{
            $\mathcal{X}' \leftarrow \mathcal{X}' \cup \{\mathcal{\mathcal{X}}_{imask}[z]\}$\\
            $\mathcal{Y}' \leftarrow \mathcal{Y}' \cup \{\mathcal{Y}[k]\}$\\
            }
        }
        
    }
 \Return $\mathcal{X'}, \mathcal{Y'}$
 \caption{$\textbf{EDDA}_{ml}$: Explanation-driven Data Augmentation for Multi-label Classification}
 \label{algor2}
\end{algorithm}
\end{frame}{}

\section{Training Settings} 
\paragraph{Train, Validation \& Test Set} 
For both CIFAR-100 dataset \cite{krizhevsky2009learning} and PASCAL VOC 2012 dataset \cite{pascal-voc-2012}, we randomly split out 20\% of training data into the validation set, making the train v.s. validation ratio 4:1. We maintained the same ratio for the training of baselines and EDDA. The validation sets are used for model-related hyperparameter tuning and the evaluation are performed on the test set.

\paragraph{Data Pre-processing} To standardize the model inputs, we performed different kinds of data pre-processing on different datasets.
\begin{itemize}
\item \textbf{CIFAR-100}: We used standard image processing techniques such as cropping and flipping following the setting in \cite{zhang2017mixup,yun2019cutmix}.
\item \textbf{PASCAL VOC 2012}: Images were resized to 224*224 and no other pre-processing are done.
\end{itemize}

\paragraph{Hyperparameters} For fair comparisons, we used the same batch size, number of epochs, learning rate and weight decay for each Dataset \& Model combination, regardless of which training methods were used. Table \ref{tab:trainsetting} includes all the non-model-related hyperparameters that we empirically chose.

\begin{table}[ht]
\caption{Hyperparameter settings of different Dataset \& Model combinations}
\centering
    \begin{tabular}{ |p{2.7cm}|p{2cm}|p{2cm}|p{2cm}|p{2cm}|}
 \hline
    Dataset & \multicolumn{2}{|c|}{CIFAR-100}    &\multicolumn{2}{|c|}{PASCAL VOC 2012}\\
 \hline
    Model & \multicolumn{1}{|c|}{ResNet-50}    &\multicolumn{1}{|c|}{VGG-16}
    & \multicolumn{1}{|c|}{ResNet-50}    &\multicolumn{1}{|c|}{VGG-16}\\
 \hline
  Batch Size &$512$ &$512$ &$8$ &$8$\\
  Number of Epochs &$200$ &$200$ &$20$ &$20$\\
  Learning Rate &$0.01$ &$0.01$ &$0.01$ &$0.01$ \\
  Weight Decay &$0.0001$ &$0.0001$ &$0.0001$ &$0.0001$ \\
 \hline
\end{tabular}
\label{tab:trainsetting}
\end{table}

Additionally, in terms of MixUp \cite{zhang2017mixup} and AugMix \cite{hendrycks2020augmix} augmentation methods, we considered the effect of parameter  \emph{$\alpha$} inside the Beta distribution in the set of {0.5, 1}, as well as the Jensen-Shannon Consistency Loss respectively. From the validation results, we empirically found that \emph{$\alpha$} did not influence the performance a lot, therefore we chose the best \emph{$\alpha$} for each random seed. As for the Jensen-Shannon Consistency Loss, it was helpful with CIFAR-100 training but would harm the PASCAL VOC 2012 a lot, and we chose not to use it as a result.

Last but not least, we tuned the \emph{imask} and \emph{umask} thresholds \emph{$\tau_{imask}$},\emph{$\tau_{umask}$} and the epoch rate \emph{r} when we got the explainer involved (Exp\_epoch\_rate) (e.g.: if the number of epochs is 200 and \emph{r} is 0.5, then the first 100 epochs would do no augmentation, and the next 100 epochs would be EDDA training). 
We tuned \emph{$\tau_{imask}$},\emph{$\tau_{umask}$} in the set of $\{0.1, 0.2, 0.3, 0.4, 0.5\}$. We empirically set the (\emph{$\tau_{imask}$},  \emph{$\tau_{umask}$}) to be (0.5, 0.5) for ResNet-50 CIFAR-100 models, and (0.2, 0.2) for VGG-16 CIFAR-100 models. Since we only use \emph{imask} for multi-label PASCAL VOC 2012 dataset, the \emph{$\tau_{imask}$} was empirically set to 0.5 for both ResNet-50 and VGG-16 models.

As for Exp\_epoch\_rate \emph{r}, we empirically found that an \emph{r} of 0.0 (i.e.: when the explainer was involved from the very beginning) led to the best Drop\% and Increase\%, while all Exp\_epoch\_rates shared similar accuracy. Therefore, 0.0 was the best \emph{r} we had chosen.

\section{Additional Experimental Results} 
Apart from Drop\% and Increase\%, we also evaluated the EDDA models as well as the baseline models with Deletion\textsubscript{zero}, Deletion\textsubscript{blur}, Insertion\textsubscript{zero} and Insertion\textsubscript{blur}, following previous works on Deletion game and Insertion game based on the change in the confidence score \cite{petsiuk2018rise}. We measured the AUC of confidence scores based on softmax probability after class logits and prediction accuracy when the image is gradually masked or recovered from mask. Regarding the type of masking, we have tried masking with blur and masking with zero (a black masking). The results of Deletion and Insertion of both types of masking are shown in Table \ref{tab:deletionInsertionZeroScore} and Table \ref{tab:deletionInsertionBlurScore}. We have good Insertion results for both zero and blur masking, and we do not perform well in Deletion, which is within our expectation, as we are exactly training on the masked images, improving the robustness on partial image prediction and hence resulting in a high confidence score even on masked images.

\begin{table}[h]
\caption{ResNet-50 and VGG-16 model performance in terms of Deletion\textsubscript{zero-score} and Insertion\textsubscript{zero-score} on CIFAR-100 and PASACL VOC 2012.}
    \centering
    
\begin{subtable}{\textwidth}
 \begin{tabular}{ |p{1.26cm}|p{2.7cm}|p{2.7cm}|p{2.7cm}|p{2.7cm}|  }
 \hline
    Model & \multicolumn{2}{|c|}{ResNet-50}    &\multicolumn{2}{|c|}{VGG-16} \\
 \hline
    Metric&  \multicolumn{1}{|c|}{Deletion\textsubscript{zero-score}}  &  \multicolumn{1}{|c|}{Insertion\textsubscript{zero-score}}   & \multicolumn{1}{|c|}{Deletion\textsubscript{zero-score}}   &\multicolumn{1}{|c|}{Insertion\textsubscript{zero-score}}  \\
 \hline
  No Aug &$22.36 \pm 0.50$ &$27.30 \pm 0.41$ &\textbf{22.58} $\pm$ \textbf{0.40} &$36.30 \pm 0.61$ \\
  CutMix &$19.83 \pm 0.51$ &$25.91 \pm 0.55$ &$25.76 \pm 0.71$ &$29.84 \pm 0.95$ \\
  MixUp &\textbf{17.53} $\pm$ \textbf{0.43} &$20.32 \pm 0.44$ &$22.71 \pm 0.65$ &$24.66 \pm 0.75$ \\
  AugMix &$23.17 \pm 0.31$ &$28.20 \pm 0.18$ &$23.47 \pm 0.76$ &$37.08 \pm 0.73$ \\
  RandExp &$22.78 \pm 0.24$ &$26.76 \pm 0.40$ &$26.57 \pm 0.55$ &$34.54 \pm 0.58$ \\
  EDDA\textsubscript{GC} &$39.77 \pm 0.50$ &\textbf{46.08} $\pm$ \textbf{0.38} &$42.52 \pm 0.80$ &\textbf{53.34} $\pm$ \textbf{0.32} \\
 \hline
\end{tabular}
    \caption{CIFAR-100}
 \end{subtable}

\begin{subtable}{\textwidth}
 \begin{tabular}{ |p{1.26cm}|p{2.7cm}|p{2.7cm}|p{2.7cm}|p{2.7cm}|  }
 \hline
    Model & \multicolumn{2}{|c|}{ResNet-50}    &\multicolumn{2}{|c|}{VGG-16} \\
 \hline
    Metric&  \multicolumn{1}{|c|}{Deletion\textsubscript{zero-score}}  &  \multicolumn{1}{|c|}{Insertion\textsubscript{zero-score}}   & \multicolumn{1}{|c|}{Deletion\textsubscript{zero-score}}   &\multicolumn{1}{|c|}{Insertion\textsubscript{zero-score}}  \\
 \hline
  No Aug &$27.16 \pm 1.16$ &$78.25 \pm 1.33$ &$23.01 \pm 0.89$ &$77.53 \pm 1.08$ \\
  CutMix &$31.86 \pm 1.14$ &$69.17 \pm 1.50$ &$27.36 \pm 0.59$ &$66.56 \pm 0.99$ \\
  MixUp &$29.82 \pm 1.01$ &$71.67 \pm 1.15$ &$23.39 \pm 0.28$ &$67.06 \pm 0.53$  \\
  AugMix &$26.15 \pm 1.15$ &$74.08 \pm 1.21$ &\textbf{19.69} $\pm$ \textbf{0.97} &$77.08 \pm 0.95$ \\
  RandExp &\textbf{21.58} $\pm$ \textbf{2.25} &$63.52 \pm 3.79$ &$23.25 \pm 0.93$ &$75.04 \pm 0.80$ \\
  EDDA\textsubscript{GC} &$34.44 \pm 4.41$ &\textbf{79.45} $\pm$ \textbf{3.23} &$48.23 \pm 2.25$ &\textbf{92.06} $\pm$ \textbf{2.01} \\
 \hline
\end{tabular}
    \caption{PASCAL VOC 2012}
 \end{subtable}

\label{tab:deletionInsertionZeroScore}
\end{table}

\begin{table}[h]
 \caption{ResNet-50 and VGG-16 model performance in terms of Deletion\textsubscript{blur-score} and Insertion\textsubscript{blur-score} on CIFAR-100 and PASACL VOC 2012.}
    \centering
    
\begin{subtable}{\textwidth}
 \begin{tabular}{ |p{1.26cm}|p{2.7cm}|p{2.7cm}|p{2.7cm}|p{2.7cm}|  }
 \hline
    Model & \multicolumn{2}{|c|}{ResNet-50}    &\multicolumn{2}{|c|}{VGG-16} \\
 \hline
    Metric&  \multicolumn{1}{|c|}{Deletion\textsubscript{blur-score}}  &  \multicolumn{1}{|c|}{Insertion\textsubscript{blur-score}}   & \multicolumn{1}{|c|}{Deletion\textsubscript{blur-score}}   &\multicolumn{1}{|c|}{Insertion\textsubscript{blur-score}}  \\
 \hline
  No Aug &$32.63 \pm 0.67$ &$39.64 \pm 0.31$ &$31.17 \pm 0.45$ &$45.60 \pm 0.64$ \\
  CutMix &$27.89 \pm 0.67$ &$36.37 \pm 0.74$ &$33.83 \pm 0.94$ &$39.47 \pm 1.01$ \\
  MixUp &\textbf{25.80} $\pm$ \textbf{0.83} &$29.55 \pm 0.98$ &\textbf{28.98} $\pm$ \textbf{0.71} &$31.76 \pm 0.97$ \\
  AugMix &$34.27 \pm 0.36$ &$40.35 \pm 0.29$ &$32.55 \pm 0.46$ &$45.66 \pm 0.91$ \\
  RandExp &$36.24 \pm 0.86$ &$42.24 \pm 0.68$ &$36.75 \pm 0.64$ &$45.10 \pm 0.75$ \\
  EDDA\textsubscript{GC} &$40.28 \pm 0.66$ &\textbf{47.32} $\pm$ \textbf{0.49} &$38.51 \pm 1.03$ &\textbf{48.85} $\pm$ \textbf{0.52} \\
 \hline
\end{tabular}
    \caption{CIFAR-100}
 \end{subtable}

\begin{subtable}{\textwidth}
 \begin{tabular}{ |p{1.26cm}|p{2.7cm}|p{2.7cm}|p{2.7cm}|p{2.7cm}|  }
 \hline
    Model & \multicolumn{2}{|c|}{ResNet-50}    &\multicolumn{2}{|c|}{VGG-16} \\
 \hline
    Metric&  \multicolumn{1}{|c|}{Deletion\textsubscript{blur-score}}  &  \multicolumn{1}{|c|}{Insertion\textsubscript{blur-score}}   & \multicolumn{1}{|c|}{Deletion\textsubscript{blur-score}}   &\multicolumn{1}{|c|}{Insertion\textsubscript{blur-score}}  \\
 \hline
  No Aug &$56.03 \pm 1.14$ &\textbf{89.89} $\pm$ \textbf{0.30} &$38.90 \pm 0.64$ &$90.98 \pm 0.23$ \\
  CutMix &$53.73 \pm 1.93$ &$81.06 \pm 1.08$ &$39.84 \pm 0.43$ &$80.10 \pm 0.24$ \\
  MixUp &$49.14 \pm 0.86$ &$82.18 \pm 0.80$ &$36.23 \pm 0.45$ &$81.15 \pm 0.45$ \\
  AugMix &$60.17 \pm 1.54$ &$85.93 \pm 0.53$ &\textbf{31.66} $\pm$ \textbf{0.62} &$91.57 \pm 0.27$ \\
  RandExp &$43.68 \pm 6.51$ &$78.76 \pm 3.96$ &$45.31 \pm 0.89$ &\textbf{93.22} $\pm$ \textbf{0.41} \\
  EDDA\textsubscript{GC} &\textbf{38.53} $\pm$ \textbf{2.11} &$82.81 \pm 2.92$ &$44.02 \pm 1.19$ &$91.64 \pm 0.89$ \\
 \hline
\end{tabular}
    \caption{PASCAL VOC 2012}
 \end{subtable}

\label{tab:deletionInsertionBlurScore}
\end{table}

\section{Criticisms and Limitations}
\begin{table}[h]
\caption{Test accuracy of ResNet-50, VGG-16 on CIFAR-100 and PASCAL VOC 2012. }
\centering
    \begin{tabular}{ |p{1.26cm}|p{2.7cm}|p{2.7cm}|p{2.7cm}|p{2.7cm}|}
 \hline
    Dataset & \multicolumn{2}{|c|}{CIFAR-100}    &\multicolumn{2}{|c|}{PASCAL VOC 2012}\\
 \hline
    Model & \multicolumn{1}{|c|}{ResNet-50}    &\multicolumn{1}{|c|}{VGG-16}
    & \multicolumn{1}{|c|}{ResNet-50}    &\multicolumn{1}{|c|}{VGG-16}\\
 \hline
  No Aug &$63.33 \pm 0.21$ &$69.65 \pm 0.38$ &\textbf{77.51} $\pm$ \textbf{0.26} &$73.95 \pm 0.26$ \\
  CutMix &\textbf{67.22} $\pm$ \textbf{0.42} &\textbf{73.43} $\pm$ \textbf{0.20} &$76.09 \pm 0.61$ &$72.12 \pm 0.22$ \\
  MixUp &$65.88 \pm 0.24$ &$71.80 \pm 0.47$ &$75.82 \pm 0.35$ &$72.56 \pm 0.40$ \\
  AugMix &$62.24 \pm 0.47$ &$69.44 \pm 0.22$ &$76.46 \pm 0.40$ &$73.99 \pm 0.19$ \\
  RandExp &$57.28 \pm 0.40$ &$66.74 \pm 0.48$ &$70.44 \pm 2.13$ &$72.72 \pm 0.23$ \\
  EDDA &$61.21 \pm 0.34$ &$69.43 \pm 0.35$ &$69.42 \pm 1.72$ &\textbf{74.69} $\pm$ \textbf{0.52} \\
 \hline
\end{tabular}
\vspace*{3mm}
\label{tab:accuracy}

\end{table}

As shown in Table \ref{tab:accuracy}, the EDDA models does not have great accuracy performance. If a model has simple structure, it would be easy to explain but it's accuracy may be sacrificed due to its simplicity. In contrast, the accuracy of a complex model can be high, but it may lead to low explanation results \cite{Murdoch2019InterpretableML}. From another perspective, the EDDA method is designed primarily to improve the explanation faithfulness, therefore it is reasonable to result in an acceptable degree of drop in accuracy, as there could be trade-offs between model and explanation faithfulness and accuracy. 

Another limitation is the lack of appropriate baselines. To the best of our knowledge, there are no explanation-driven data augmentation methods that are also targeting to improving explanation faithfulness in the training pipeline. As a result, we could only compare with non-explanation-driven data augmentation methods and the trivial Random Explainer baseline.

\newpage

\end{document}